\documentclass[11pt]{article}
\PassOptionsToPackage{table}{xcolor}
\usepackage[preprint]{acl} 

\usepackage{times}
\usepackage{latexsym}
\usepackage[T1]{fontenc}
\usepackage[utf8]{inputenc}
\usepackage{microtype}
\usepackage{inconsolata}
\usepackage{booktabs}
\usepackage{amsmath}
\usepackage{graphicx}
\usepackage{multirow}
\usepackage{xspace}
\usepackage{subcaption}
\usepackage{enumitem}
\usepackage{makecell}
\usepackage{placeins}
\usepackage[most]{tcolorbox}

\newcommand{\wanda}{\textsc{Wanda}\xspace}

\newcommand{\qk}{Qwen3-8B\xspace}
\newcommand{\mybox}[1]{
\begin{tcolorbox}[boxrule=0pt,frame hidden,sharp corners,enhanced,borderline west={2pt}{0pt}{black}]
#1
\end{tcolorbox}
}

\title{The Benchmark Illusion: \\ Pruned LLMs Can Pass Multiple Choice but Fail to Answer}

\author{ 
 \textbf{Rui Wen\textsuperscript{1}},
 \textbf{Lu Sun\textsuperscript{2}},
 \textbf{Jiayang Liu\textsuperscript{3}},
 \textbf{Zesheng Xu\textsuperscript{4}},
 \textbf{Tianshuo Cong\textsuperscript{5}},
 \textbf{Zheng Li\textsuperscript{5}}
\\
\\
 \textsuperscript{1}Institute of Science Tokyo,
 \textsuperscript{2}Tohoku University,
 \textsuperscript{3}Nanyang Technological University
\\
 \textsuperscript{4}KTH Royal Institute of Technology,
 \textsuperscript{5}Shandong University
 }

\begin{document}
\maketitle

\begin{abstract}

Compressing large language models reduces memory use and inference cost, but it can also create failures that standard benchmarks miss. A pruned model may still perform well on multiple-choice evaluations, yet fail to answer the same question in open generation. We ask what pruning changes: does it erase the correct answer, or does it make the answer harder to produce as the top output?

We study this question with multilingual question answering, tracking the same questions before and after pruning. We find a benchmark illusion. Under high-sparsity pruning, especially \wanda, models often fail in greedy open generation while still selecting the correct answer under multiple-choice scoring. In these recognition-only errors, the answer is usually not gone, but demoted: it often reappears with beam search, sampling, or one in-context example. Overall, multiple-choice benchmarks can overstate the usability of compressed LLMs, creating an evaluation blind spot. Compressed models should be tested on what they can produce, not only on what they can recognize.
\end{abstract}
  
\section{Introduction}
\label{sec:intro}

Large language models are expensive to serve. To cut costs, developers often compress them using techniques like pruning~\citep{FC19,DLTLPWC24,FYMHPKMW24}, quantization~\citep{FSA22,DLBZ22,DPHZ23}, and other techniques~\citep{ACNHH24}, relying on standard multiple-choice benchmarks to prove the model's performance remains intact. While these benchmark scores are helpful, they hide a major practical flaw: a compressed model might ace a multiple-choice test but completely fail to answer that exact same question when a user asks it directly.

This paper investigates that gap. When a compressed model gives a wrong answer, one of two things has happened: either the model has completely lost the knowledge, or the knowledge is still there but just isn't strong enough to be the model's top response. When the latter happens, benchmark leaderboards can look healthy, but actual open-ended answering degrades. We call this the \emph{benchmark illusion}.

Our main tool is a paired-item test. We evaluate the \emph{same question} in two formats: open generation, where the model must produce the answer from scratch, and multiple-choice scoring, where the correct answer appears among candidates. We first keep only questions that the uncompressed model answers correctly in both formats. We then prune the model and ask: does the answer fail in both formats, or does the model still select it from candidates while failing to generate it?

We used multilingual question answering (QA) as our stress test because it naturally creates a wide range of pruning errors to study. For example, when applying aggressive compression (high-sparsity \wanda \cite{SLBK24}) to a Qwen \cite{T252} model, English performance barely drops, but languages like Swahili suffer massive losses in open generation. This gives us a wealth of test cases where the original model was correct, but the compressed model failed.

Results show that the answer is often demoted, not erased. Many answers that the original model could generate become ``recognition-only'' after pruning. The compressed model fails to generate the answer on its own, but still selects it from a multiple-choice list. The correct first token has not disappeared; it often drops from rank 1 to around rank 3 or 4, and can often be recovered with broader decoding methods such as beam search or sampling, or with a single in-context example. We replicate this pattern across different model families and datasets, while also identifying one key boundary: smaller Qwen models more often lose the underlying recognition ability itself.

We make three contributions:
\begin{itemize}
  \item \textbf{A New Diagnostic Tool:} We introduce a paired-item test that evaluates compressed models by comparing their ability to generate an answer, recognize it from a list, or find it through broader search methods.

  \item \textbf{Exposing the Multiple-Choice Illusion:} We demonstrate that aggressive pruning frequently destroys a model's ability to greedily generate an answer, even while its multiple-choice recognition remains perfectly intact. We refer to these as ``MC-only'' errors.

  \item \textbf{Explaining the Mechanics:} We show that the answer is often demoted to a lower rank rather than deleted, and that this demotion can often be reversed with broader decoding or light prompting. We also show when this interpretation breaks down: smaller Qwen checkpoints more often lose recognition itself.
\end{itemize}

Ultimately, a retained benchmark score does not guarantee retained usability. To understand how compressed LLMs behave in real use, we must test what they can actually produce, not only what they can recognize.

\section{Diagnostic Framework}
\label{sec:framework}

To understand how compression breaks a model, we need to distinguish recognizing an answer from producing it. A compressed model may pass a multiple-choice item because the benchmark supplies the correct answer as a candidate, but still fail when the model must answer directly. This failure mode is plausible under one-shot pruning methods such as \wanda~\citep{SLBK24} and SparseGPT~\citep{FA23}, which score and remove weights without retraining the model from scratch. Pruning may keep the answer available under some probes, but it is no longer strong enough to become the model's top output.

\subsection{The Generation Gap}
\label{sec:generation_gap}

To measure this gap, we test the model across three levels of answer access. 
\begin{itemize}
    \item \textbf{Spontaneous generation} is the hardest setting: the model must generate the answer greedily, without seeing any candidates.
    \item \textbf{Prompted recognition} is the multiple-choice test. The model is given a list of options and simply has to select the correct one. We use ``recognition'' loosely here, as providing multiple-choice options inherently gives the model formatting clues. However, it effectively measures whether the model can still succeed when a benchmark hands it the answer.
    \item \textbf{Reachability} asks whether the answer appears under broader decoding, such as beam search or sampling, even if greedy decoding fails.
\end{itemize}

We quantify the gap between generation and recognition with:
\begin{equation}
    \Delta_\text{PR} = \hat{d}_\text{FF} - \hat{d}_\text{MC},
    \label{eq:prgap}
\end{equation}
Here, $\hat{d}$ is the headroom-normalized accuracy drop after pruning. We use FF to denote free-form open generation and MC to denote multiple-choice scoring. A positive $\Delta_\text{PR}$ means that open generation degrades more than multiple-choice recognition. A negative value means that recognition degrades as much as, or more than, generation. If compression caused only a uniform drop in model quality, this gap would stay close to zero.

\subsection{Multilingual QA as a Stress Test}
\label{sec:setup}

To test whether this gap appears in practice, we need a controlled setting that produces enough pruning errors to analyze. We use multilingual question answering (QA) as our stress test, with \qk~\citep{T252} evaluated on TyDiQA~\citep{CCCGKNP20}.

The goal is not to study multilinguality itself. We use multilingual QA because a single pruning pass damages different languages by very different amounts while keeping the task and compression procedure fixed. For example, under high-sparsity \wanda \citep{SLBK24} at 0.5 sparsity, English open-generation accuracy drops by only 1.1 percentage points, while Swahili drops by 21.3 points. This variation gives us many cases where the original model answers correctly but the pruned model fails, all under one controlled compression setting.

For a true apples-to-apples comparison, we run a paired analysis. We take a fixed set of 200 questions per language and evaluate them across three formats using the exact same prompt context:
\begin{itemize}
    \item \textbf{open generation}, the model answers greedily without seeing any choices.
    \item \textbf{candidate-shown generation}, the prompt lists the correct answer together with three distractors, and the model must generate the answer. 
    \item \textbf{multiple-choice scoring}, we score the four answer options by log-likelihood and choose the highest-scoring option.
\end{itemize}

Finally, to ensure this ``benchmark illusion'' is not specific to this setup. We replicate the pattern on a second dataset, XQuAD~\citep{ARY20}; on different model families, including Mistral-7B-Instruct~\citep{JSMBCCBLLSLLSSLWLS23} and Phi-3-mini~\citep{M24}; across Qwen3~\citep{T252} model sizes; and under different calibration settings. As we show below, it is a persistent feature of model compression.

\section{Demotion, Not Erasure}
\label{sec:production}

We now address the core question: when pruning causes a model to answer incorrectly, is the knowledge completely erased, or does it merely fail to generate it?
These two hypotheses make very different predictions. If pruning completely removes the answer from the model, it should fail to select it from a multiple-choice list, push the answer far down the probability distribution, and rarely produce it under broader decoding methods like beam search. Conversely, if pruning merely demotes the answer, the model should still recognize it, rank it near the top of the distribution, or successfully generate it when decoding is less greedy.

\begin{center}
{\setlength{\fboxsep}{6pt}
\fbox{\begin{minipage}{0.90\linewidth}
\textbf{Running example.}
For one test item (``When were the Azores discovered?''), the uncompressed model correctly generates ``Around 1427''. After pruning, the model's greedy output shifts to the incorrect answer ``the 15th century''. However, the first token of the correct answer still ranks 3rd in the model's internal probability distribution, and the pruned model successfully selects the exact correct option during multiple-choice scoring. This perfectly illustrates the failure mode we are measuring: the answer is demoted, not erased.
\end{minipage}}}
\end{center}

\subsection{The Paired Paradox}

We first test the most direct prediction of demotion: pruning should sometimes break open generation while leaving multiple-choice recognition intact. The clearest evidence comes from comparing the same questions before and after pruning.

We isolate items that the uncompressed model answers correctly in both open generation and multiple-choice scoring. We then ask: when pruning hurts the model's ability to generate the answer, does its ability to select the answer from candidates survive? We measure this with the dissociation rate: the fraction of both-correct items that become multiple-choice-only (MC-only) after pruning.

\begin{figure}[t]
\centering
\includegraphics[width=0.9\linewidth]{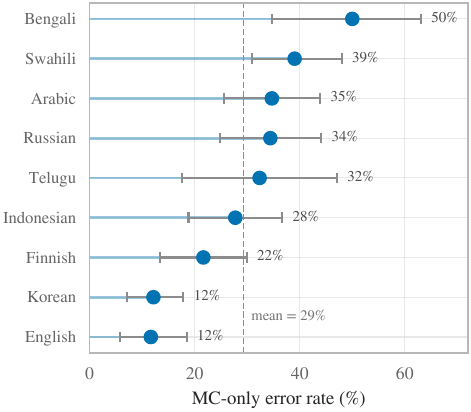}
\caption{\textbf{Many producible answers become recognition-only after pruning.} The pruned model fails in greedy open generation but still selects the gold answer from candidates. The full four-way fate breakdown is shown in Figure~\ref{fig:fatestacked}.}
\label{fig:fate}
\end{figure}

Figure~\ref{fig:fate} shows that recognition often survives. Among items that the model could previously both generate and recognize, pruning often breaks only the generation behavior. The pruned model fails to generate the answer greedily, but still selects it from the candidate list. This dissociation occurs for 39\% of Swahili items, 35\% of Arabic items, and 34\% of Russian items, compared with 12\% for English. Bengali reaches 50\%. Because all of these items were generated and recognized before pruning, a switch to MC-only cannot be explained by a complete loss of candidate-supported recognition. The answer remains selectable when it is supplied as an option, even though it no longer appears as the greedy open-generation output.

\mybox{
\textbf{Takeaway.} Pruning frequently breaks greedy answer production before it breaks candidate-supported recognition.}

\subsection{More Support Means Less Damage}

The paired analysis above gives item-level evidence: pruning can break greedy answer production while leaving candidate-supported recognition intact. We next ask whether the same pattern appears in overall accuracy trends. Here, ``overall'' means that we no longer track each item's fate separately; instead, we average the pruning drop for each evaluation format across the same items and languages.

To do this, we compare an \emph{output-mode ladder}: a sequence of evaluation formats that give the model increasing amounts of answer support. The ladder has three steps. In open generation, the model receives no answer candidates and must generate the answer greedily. In candidate-shown generation, the prompt lists the correct answer together with three distractors, and the model must generate the answer. In multiple-choice scoring, the model does not generate freely; instead, each candidate answer is scored by log-likelihood, and the highest-scoring option is selected.

\begin{figure}[t]
  \centering
  \includegraphics[width=0.88\linewidth]{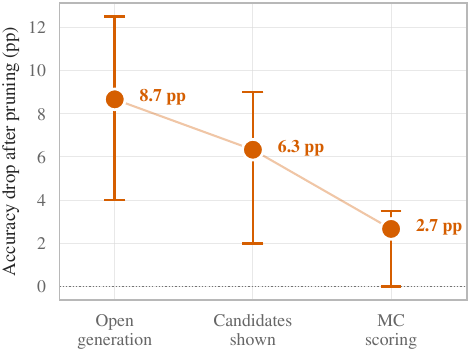}
  \caption{\textbf{More answer support means a smaller pruning drop.}
    Open generation drops most and MC scoring least.}
  \label{fig:ladder}
\end{figure}

This ladder tests a simple prediction. If pruning mainly makes answers harder to produce, then formats with more answer support should suffer less damage, which is what we observe in Figure~\ref{fig:ladder}. On average, open generation drops the most, multiple-choice scoring drops the least, and candidate-shown generation falls in between.

\paragraph{Do candidates help only when they contain the answer?}
A natural concern is that candidate lists may help for a trivial reason: they show the model the expected response format, even if the correct answer is absent. We test this with two ablations of the candidate-shown condition. In \emph{distractors-only}, the prompt shows the three distractors but omits the gold answer. In \emph{unrelated}, the prompt shows four unrelated candidates.

These ablations do not improve generation. Accuracy is 46.9\% with no candidates, 42.2\% with distractors-only, and 32.1\% with unrelated candidates. By contrast, when the prompt includes the gold answer together with the distractors, accuracy rises to 75.8\%. Thus, candidate lists help only when they contain the correct answer. The gain therefore reflects the model using the offered answer, not merely benefiting from candidate formatting.

\mybox{
\textbf{Takeaway.} As the evaluation format gives the model more direct access to the answer, pruning damage becomes smaller; this supports the view that compression hurts answer production more than answer selection.}

\begin{figure}[t]
  \centering
  \includegraphics[width=0.74\linewidth]{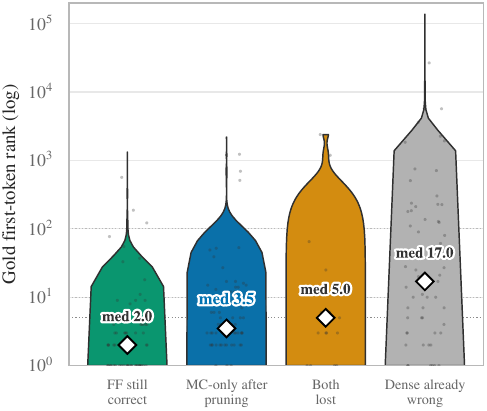}
  \caption{\textbf{Pruning demotes gold answers rather than erasing them}.
    On both-correct items, recognition-only answers stay near the top of the
    next-token distribution after pruning: the gold first token has a median rank of 3.5, which is close enough to remain reachable but no longer selected by greedy decoding.}
  \label{fig:goldsignal}
\end{figure}

\begin{figure*}[t]
  \centering
  \begin{subfigure}[t]{0.41\textwidth}
    \centering
    \includegraphics[width=\linewidth]{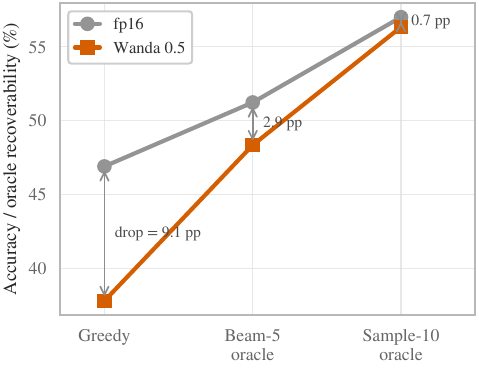}
    \caption{Mean greedy accuracy vs.\ oracle recoverability}\label{fig:decoding_a}
  \end{subfigure}\hfill
  \begin{subfigure}[t]{0.41\textwidth}
    \centering
    \includegraphics[width=\linewidth]{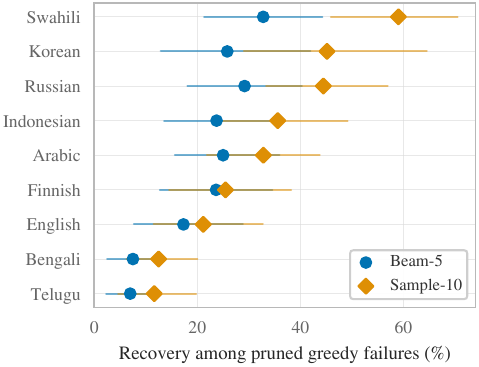}
    \caption{Beam-5 and 10-sample recovery by language}\label{fig:decoding_b}
  \end{subfigure}
  \caption{\textbf{Missed answers remain reachable from the distribution}.
    Pruning lowers greedy top-1 accuracy but barely changes whether the answer
    appears under broader decoding. Among items where pruned greedy decoding fails,
    beam-5 and 10-sample recovery is substantial for several languages but low for
    low resource languages.}
  \label{fig:decoding}
\end{figure*}

\subsection{Where Did the Answer Go?}

The previous sections show that many pruned failures remain correct under multiple-choice scoring. But this does not yet explain where the answer is inside the model. One possibility is that the candidate format simply makes the task easy by showing the answer. Another possibility is that the answer is still close to being produced, but no longer wins greedy decoding. To distinguish these cases, we inspect the model's output distribution directly.

We first measure the rank of the gold answer's first token. This directly tests whether the answer has disappeared or merely lost the top-token competition. Under greedy decoding, only the rank-1 token is selected; a token at rank 3 or 4 is close to being produced, but still loses to a few competitors. We compute this rank under fp16 and under \wanda sp\,=\,0.5, as shown in Figure~\ref{fig:goldsignal}. The key result comes from the MC-only items: when pruning breaks open generation but preserves multiple-choice correctness, the gold first token has a median rank of 3.5 after pruning. By contrast, items that the uncompressed model already fails have a median rank of 17. Thus, MC-only answers are usually near misses: the answer remains close to the top of the distribution, but is no longer selected by greedy decoding.

The broader rank pattern supports the same interpretation. After pruning, the median gold-token rank is at most 6 for eight of nine languages on this subset (Telugu is 14). This is still within the top 0.01\% of the 151{,}936-token vocabulary, far above the rank expected for a random token. In other words, pruning usually does not bury the answer deep in the vocabulary; it demotes the answer from the winning position to a near-top alternative.

These results explain how greedy generation can fail even when the answer remains close. Pruning shifts probability mass near the top of the distribution. The gold token may fall from rank 1 to rank 3 or 4. That small shift is enough for greedy decoding to choose a competitor. Once the first token is wrong, errors can compound over the rest of the answer.
\mybox{
\textbf{Takeaway.} MC-only answers are usually near misses: pruning demotes them below a few competitors rather than pushing them out of the distribution.}

\subsection{Recovering Missed Answers}
\label{sec:decoding}

If pruning primarily demotes answers rather than erasing them, we should be able to dig those missed answers back up using broader search techniques.
\paragraph{Broader decoding recovers many missed answers.}
To test this, we compared standard greedy decoding against oracle recoverability, that is, checking whether the correct answer appears anywhere within 5 beam-search paths or 10 nucleus samples ($p=0.9$) on a fixed 100-item subset. While oracle recoverability is an upper bound rather than a standalone accuracy metric, it tells us whether the answer is still accessible in the model's internal distribution. 

Figure~\ref{fig:decoding} shows that many missed answers remain reachable. For example, Russian rises from 28\% greedy accuracy to 49\% beam-5 recoverability and 60\% 10-sample recoverability. Swahili rises from 39\% to 55\% and 74\%, and Arabic from 36\% to 49\% and 57\%. Overall, every single language gains between 3 and 21 percentage points just by using beam search.

The uncompressed fp16 model provides a crucial baseline here. While pruning causes standard greedy accuracy to drop by 9.1 percentage points (46.9\% to 37.8\%), the 10-sample recoverability rate barely changes at all, dropping a mere 0.7 percentage points (57.0\% to 56.3\%). Beam-5 recoverability also sees a minimal drop of only 2.9 percentage points. This suggests that pruning appears to damage the model's ability to select the answer as its top choice; it does not destroy the answer's reachability.

\paragraph{The recoverable failures are concentrated among pruned errors.}
The recoverability numbers above include items that greedy decoding already answers correctly. A stricter test focuses only on failures: among items where pruned greedy decoding fails, how often does broader decoding still reach the gold answer?

Among the 560 pruned greedy failures, beam-5 recovers 20.2\%, and 10-sample decoding recovers 30.2\%. These rates are roughly twice as high as for the uncompressed model's own greedy failures. The sampling confidence intervals do not overlap, and the result is stable across five random seeds: 31.4\% recovery for pruned failures versus 19.5\% for uncompressed failures.

This means pruning creates a special kind of failure: the answer is missed by greedy decoding, but is still unusually easy to recover from the distribution. The effect is not uniform across languages. Swahili greedy failures recover at 59\% under sampling, while Bengali and Telugu recover at only 12\%. This suggests a boundary: for some languages, especially those with more difficult tokenization, a larger share of the pruning loss may reflect genuine loss under our probes rather than a decoding-only failure.

\paragraph{One in-context example also helps.}
Light prompting gives another test of the same idea. If some pruning errors are near misses, then a small amount of task support should reduce the gap between the uncompressed and pruned models. We therefore add one short in-context QA example and compare 0-shot and 1-shot open generation for both fp16 and \wanda sp\,=\,0.5 on the same fixed subset. This lets us track the prune gap, defined as fp16 accuracy minus pruned accuracy, under both prompting conditions. Setup details are in Appendix~\ref{app:fewshot}.

A single example sharply reduces the prune gap for several languages. The gap falls from 12.5\% to 1.0\% for Arabic, from 23.5\% to 5.5\% for Swahili, and from 14.0\% to 3.5\% for Russian. This suggests that some pruning failures can be repaired with a small amount of prompting support, rather than reflecting complete loss under our probes. The result is consistent with the production-recognition interpretation, although the example may also help by clarifying the expected format or answer style.

\mybox{
\textbf{Takeaway.} Many pruned greedy failures can be recovered with broader decoding or light prompting, which supports the view that pruning often demotes answers rather than erasing them.}

\subsection{Is the Gap an Evaluation Artifact?}

A natural objection is that multiple-choice and open-generation accuracies start from very different baselines, so raw percentage-point drops are not directly comparable. We therefore use headroom normalization, which measures each drop relative to the accuracy that could be lost from that format. The same ordering remains: the mean normalized drop is $0.21$ for open generation, $0.10$ for candidate-shown generation, and $0.05$ for MC scoring. This suggests that the gap is not just an artifact of comparing formats with different baseline accuracies.

A second concern is that MC scoring may be easy because the distractors are too different from the gold answer. To test this, we replace the standard distractors with \emph{hard distractors}: the three other answers that are most similar to the gold answer by character trigram overlap. The result changes little. MC accuracy is $85.0\%$ in fp16 and $82.1\%$ after pruning, compared with $90.0\%$ and $86.8\%$ using standard distractors. Thus, candidate-supported scoring remains robust even when the choices are lexically confusable.

The paired result is also robust to hard distractors. On the both-correct set, the dissociation rate is essentially unchanged: $26.5\%$ with hard distractors versus $27.2\%$ with standard distractors. In other words, the MC-only pattern is not an artifact of easy answer choices.

\mybox{
\textbf{Takeaway.} The gap is not explained by different baseline accuracies or easy distractors; pruning still hurts open production more than candidate-supported recognition.}

\section{Generality and Boundaries}

\subsection{Across Models and Datasets}
\label{sec:generality}

\begin{figure}[t]
\centering
\includegraphics[width=0.92\linewidth]{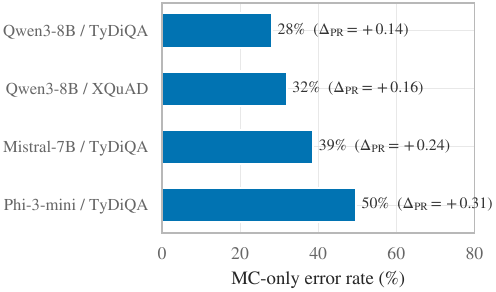}
\caption{\textbf{The dissociation generalizes across models and datasets}. For two additional model families and a second dataset (XQuAD), the recognition-only error rate remains substantial, and the production-recognition gap $\Delta_\text{PR}$ is positive in every setting. }
\label{fig:generalization}
\end{figure}

The paired analysis above uses \qk on TyDiQA as the primary testbed. We next ask whether the same dissociation appears beyond this specific model, dataset, and prompt format. Figure~\ref{fig:generalization} applies the same paired protocol to two additional model families and a second dataset.

The pattern appears in every setting we test. On TyDiQA, the recognition-only rate is 38.7\% for Mistral-7B-Instruct and 49.6\% for Phi-3-mini, both prompted with their native chat templates, compared with 28.1\% for our completion-prompted \qk. The pattern also appears on XQuAD: \qk has a recognition-only rate of 31.9\% and $\Delta_\text{PR}=+0.16$, close to its TyDiQA result of 28.1\% and $\Delta_\text{PR}=+0.14$. Across all four settings, $\Delta_\text{PR}$ is positive, meaning that open generation degrades more than multiple-choice recognition.

These results show that the benchmark illusion is not specific to one dataset or one model family. At the same time, they also caution against a universal scaling claim. Phi-3-mini shows a strong positive gap despite being smaller than the larger Qwen checkpoints. We therefore treat the Qwen scale sweep below as a within-family boundary, not as a general law about model size.

\subsection{A Qwen-Family Scale Boundary}
\label{sec:scale}

\begin{figure}[t]
\centering
\includegraphics[width=0.9\linewidth]{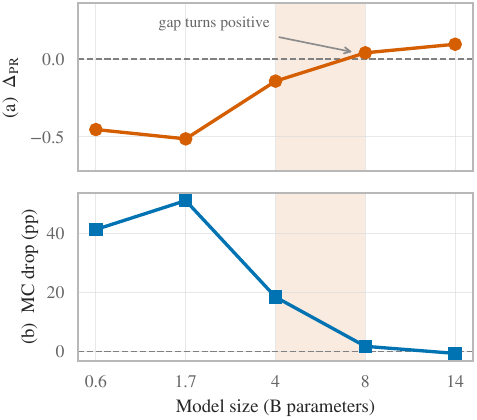}
\caption{\textbf{The Qwen-family scale sweep suggests a larger-model boundary}. \textbf{(a)} The production-recognition gap $\Delta_\text{PR}$ rises with model size and crosses zero when the model is larger than 8B. \textbf{(b)} Recognition loss shrinks with model size: smaller Qwen models lose multiple-choice accuracy too, whereas larger Qwen models mostly preserve it.}
\label{fig:scale}
\end{figure}

We next ask when the production-recognition dissociation appears within a single model family. Figure~\ref{fig:scale} sweeps Qwen3 models from 0.6B to 14B parameters. The pattern changes with scale. In smaller Qwen3 models, pruning damages multiple-choice accuracy as much as, or more than, open generation ($\Delta_\text{PR}\le 0$). This means that many answers are no longer selected even when candidates are supplied. For these models, pruning looks less like answer demotion and more like loss of knowledge of the answer.

Larger Qwen models behave differently. For the 8B and 14B checkpoints, multiple-choice accuracy is mostly preserved while open generation still degrades. The MC loss is at most 2\%, and $\Delta_\text{PR}$ becomes positive. In this regime, multiple-choice scores can look stable even as greedy answer generation gets worse.

One possible explanation is model capacity. Smaller models may have less redundancy, so pruning can remove information that is needed even for candidate-supported recognition. Larger models may store answer-relevant signals more redundantly. After pruning, the answer may therefore remain available in the distribution, but become weaker and lose the top-output competition. This interpretation is consistent with the scale sweep, but we treat it as a hypothesis: our experiments identify the behavioral boundary, not the internal mechanism that causes it.

This result gives both a warning and a boundary. The warning is that, for larger Qwen models, multiple-choice evaluation can hide failures that users would see in open answering. The boundary is that this effect does not appear equally across all sizes: only the 8B and 14B Qwen models show a clearly positive gap. We therefore treat the exact transition point as suggestive rather than definitive. 

\mybox{
\textbf{Takeaway.} In the Qwen family, the benchmark illusion appears mainly in larger checkpoints: multiple-choice accuracy can stay stable while open answering gets worse, whereas smaller checkpoints more often lose candidate-supported recognition too.}

\section{Better Calibration Helps, but the Gap Remains}
\label{sec:calibration}

One possible explanation is that the production-recognition gap is a side effect of calibrating the pruning process exclusively on English text. If the pruning mask simply discards the ``wrong'' languages, then switching to a balanced, multilingual calibration set should preferentially rescue open-ended generation and close the gap.

We find that calibration helps, but not in the way that would close the gap: it improves both open generation and multiple-choice scoring, so the production-recognition gap remains.

\begin{figure}[t]
\centering
\includegraphics[width=0.72\linewidth]{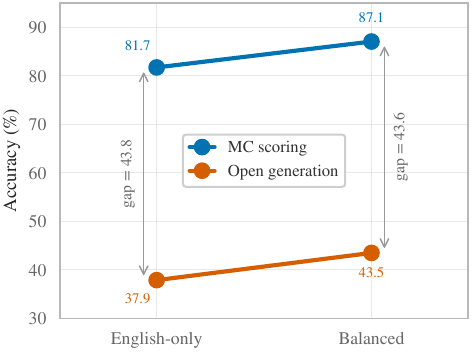}
\caption{\textbf{Calibration lifts both formats but leaves the gap.} Mean open-generation and MC accuracy under English-only vs.\ balanced multilingual calibration (9 languages). }
\label{fig:calib}
\end{figure}

Calibration clearly improves retained accuracy. Per-language calibration recovers much of the lost open-generation performance, and balanced multilingual calibration is a more practical alternative that also helps. However, calibration does not preferentially close the production-recognition gap. As shown in Figure~\ref{fig:calib}, balanced calibration raises both open-generation and multiple-choice accuracy, but the mean format gap, measured as MC$-$FF accuracy, stays almost unchanged ($43.8$ vs. $43.6$ points). 

\mybox{
\textbf{Takeaway.} Better calibration makes pruning less damaging overall, but it does not close the production-recognition gap.}

\section{Related Work}
\label{sec:related}

\paragraph{LLM pruning and multilingual degradation.}
Model pruning reduces memory and computation by removing less important parameters. Early work studied magnitude pruning and sparse training~\citep{HPTD15,FC19,LWK18}; recent LLM pruning methods focus on one-shot compression without full retraining. SparseGPT~\citep{FA23} uses layer-wise reconstruction with second-order approximations, while \wanda~\citep{SLBK24} scores weights using magnitude and activation statistics. Other work explores symbolic pruning metrics~\citep{DLTLPWC24} and optimization-based mask learning~\citep{FYMHPKMW24}. Most pruning studies evaluate compressed models with perplexity, reconstruction quality, or downstream accuracy. Recent work shows that these metrics can hide important failures: LLM-KICK~\citep{JGDZWY24} finds that perplexity can remain stable while knowledge-intensive tasks degrade, and multilingual pruning studies show that WANDA and SparseGPT can damage languages unevenly, with calibration language affecting what is preserved~\citep{CT25,KCFZ26}. Our work builds on these findings but asks a different question: when pruning causes an error, is the answer erased, or does it remain recognizable or reachable but harder to produce?

\paragraph{Benchmark format gaps.}
Evaluation format can strongly affect model behavior. Multiple-choice benchmarks are easy to score, but they do not test the same behavior as open generation. 
\citet{PPJ25} show that models may generate the correct answer in free-form settings but select the wrong option in MCQ format. \citet{RKWNY25} further show that open-ended generation and multiple-choice selection exhibit different correction dynamics and failure modes.
We study the complementary failure mode in compressed models: after pruning, a model can still select the correct answer under multiple-choice scoring while failing to generate it directly. Our paired-item diagnostic tracks the same questions before and after pruning to separate answer production, candidate-supported recognition, and broader-decoding reachability.

\section{Conclusion}
\label{sec:conclusion}

Pruned models can pass the benchmark and still fail the user. A wrong open-generation answer does not always mean that the answer is unavailable. In many cases, the answer remains recognizable, high-ranked, or reachable under broader decoding, but is no longer selected as the greedy output.

This is the benchmark illusion. Multiple-choice accuracy can miss failures that matter for user-facing generation. Evaluating compressed LLMs therefore requires asking not only whether the model can recognize an answer, but where that answer remains available: as a generated output, as a selected candidate, or only as a reachable alternative under broader decoding.

\section*{Limitations}
\label{sec:limitations}

Our notion of recognition is loose. Multiple-choice scoring does not prove that the model recognizes an answer in a cognitive sense, because answer choices provide lexical anchors and format cues. Open generation and multiple-choice scoring also differ in more than one way. We reduce this concern with paired-item analysis, candidate ablations, hard distractors, answer-rank analysis, and decoding recovery, but the production-recognition distinction is not perfectly isolated.

Our evidence is mainly behavioral and centered on high-sparsity pruning in multilingual QA. We use multilingual QA as a controlled stress test, not as the only setting where the benchmark illusion may appear. The effect may differ under other compression methods, lower sparsity levels, retraining after pruning, or other tasks such as summarization, translation, tool use, and long-form generation. We also do not identify the exact circuits or weight groups responsible for demoting answers. Mechanistic analysis and broader task coverage are important next steps.

\bibliography{normal_generated_py3}

\appendix

\section{Full Fate Breakdown}
\label{app:figtables}

\begin{figure}[h]
\centering
\includegraphics[width=\linewidth]{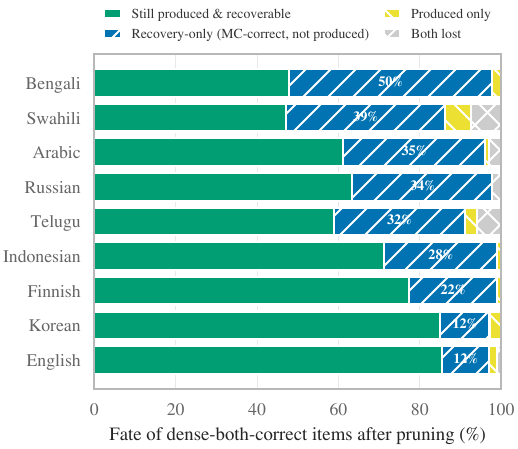}
\caption{\textbf{Full four-way fate of uncompressed-both-correct items after pruning} (\wanda sp\,=\,0.5), the breakdown summarized by Figure~\ref{fig:fate}. Each language's both-correct items are split into: still produced \& recognized, MC-only (recognized but not produced), produced-only, and lost in both formats. Few items are lost in both formats; the MC-only segment (labelled) is the dissociation rate in Figure~\ref{fig:fate}.}
\label{fig:fatestacked}
\end{figure}

\section{Reproducibility Details}
\label{app:reproducibility}

All data sampling uses \texttt{random.seed(42)}.
Calibration uses the first 128 passages from a fixed C4~\citep{RSRLNMZLL20} subset.
Per-language calibration uses 64 Wikipedia passages per language, disjoint from evaluation items.
Evaluation is custom rather than based on \texttt{lm-eval-harness}.
All experiments use bfloat16 on A6000/Ada 6000 GPUs.
Code and result JSONs will be released upon publication.

Open-generation QA (FF) uses the prompt:
\begin{quote}\small
\texttt{Context: \{passage\}\textbackslash n\textbackslash nQuestion: \{q\}\textbackslash n\textbackslash nAnswer in a few words:}
\end{quote}
with greedy decoding and \texttt{max\_new\_tokens=32}.
Scoring uses normalized substring containment after lowercasing and punctuation removal.
Multiple-choice benchmarks use log-likelihood of the answer option rather than generation unless otherwise specified.

\paragraph{Evaluation format details.}
For completeness we specify the evaluation choices that could affect the production-recognition gap:
(1) Context passages are always in the target language; prompts use the English template above for all languages (we did not translate the template, matching common practice and isolating model behavior from prompt-translation quality).
(2) MC distractors for TyDiQA~\citep{CCCGKNP20} are sampled from other gold answers in the same language, so all options are language-appropriate; the gold option is excluded from the distractor pool by normalized string match.
(3) Gold-answer matching uses normalized substring containment; we additionally report token-F1 and exact-match in Section~\ref{sec:production}, which give the same direction.
(4) For scripts without whitespace word boundaries (e.g., Thai), tokenizer fertility is computed per character rather than per word.
(5) Gold first-token rank and margin use the leading-space token variant ($\langle\text{space}\rangle$+answer), since that is what the model predicts after the colon-terminated prompt.
(6) Truncation: we cap open-generation prompts at 512 tokens and 1-shot prompts at 1024; truncation rates are near zero for 0-shot and reported per language for few-shot (Section~\ref{sec:production}).
(7) \texttt{max\_new\_tokens=32} is sufficient for short-answer QA across all languages; we strip any Qwen3~\citep{T252} \texttt{<think>} spans before matching.

\paragraph{Subset consistency.}
The core Section~\ref{sec:production} analyses (open generation, MC, cued generation, 1-shot, and the paired dissociation) use a single fixed 200-item-per-language subset, selected with an independent per-language \texttt{random.Random(42)} shuffle; the beam/sampling decoding runs use the first 100 items of the same subset.
The gold-answer rank and category-level full-sequence log-probability in Section~\ref{sec:production} use the same fixed 200-item-per-language subset as the paired analysis.
For all rank analyses, we use the in-context leading-space gold token and verify that the joint-encoding boundary token matches the answer's first token, including for non-Latin scripts.
Reporting the open-generation and candidate-supported formats on the same items removes cross-subset discrepancies and supports the paired item-level analysis.

\section{Few-Shot Setup}
\label{app:fewshot}

The 1-shot results in Section~\ref{sec:production} use one in-context exemplar drawn from the same language but disjoint from the evaluation subset (the lowest-index item not in the eval set; the same exemplar is used for all eval items of a language).
The exemplar context is truncated to 300 characters and the evaluation context to 500 characters to limit prompt length; prompts are capped at 1024 tokens.
The 0-shot baseline in Table~\ref{tab:fewshot} uses the same 500-character evaluation-context truncation as the 1-shot condition, so the only difference between the two columns is the presence of the exemplar.
The exemplar follows the format \texttt{Context: \{c\}\textbackslash nQuestion: \{q\}\textbackslash nAnswer in a few words: \{a\}}, followed by a blank line and the evaluation item in the same format with the answer omitted.
Earlier drafts reported few-shot results on a separate subset with a longer (5-shot) prompt that truncated for high-fertility scripts; the unified 1-shot setup avoids this and is reported throughout.

\begin{table}[h]
\centering\small
\caption{Open-generation accuracy (\%) at 0-shot and 1-shot for fp16 and \wanda sp\,=\,0.5, and the prune gap (fp16\,$-$\,W0.5) at each, on the fixed subset (summarized in Section~\ref{sec:production}). One exemplar sharply reduces the prune gap for most languages.}
\label{tab:fewshot}
\begin{tabular}{l@{\hskip 5pt}rr@{\hskip 7pt}rr@{\hskip 7pt}rr}
\toprule
& \multicolumn{2}{c}{0-shot} & \multicolumn{2}{c}{1-shot} & \multicolumn{2}{c}{prune gap} \\
\cmidrule(lr){2-3}\cmidrule(lr){4-5}\cmidrule(lr){6-7}
Lang & fp16 & W.5 & fp16 & W.5 & 0-sh & 1-sh \\
\midrule
arabic  & 49.0 & 36.5 & 66.5 & 65.5 & 12.5 & 1.0 \\
russian & 47.5 & 33.5 & 56.0 & 52.5 & 14.0 & 3.5 \\
swahili & 63.0 & 39.5 & 57.0 & 51.5 & 23.5 & 5.5 \\
bengali & 25.5 & 15.5 & 64.0 & 56.5 & 10.0 & 7.5 \\
telugu  & 18.0 & 12.5 & 24.5 & 23.5 &  5.5 & 1.0 \\
\bottomrule
\end{tabular}
\end{table}

\end{document}